\documentclass[preprint]{article}
\usepackage[T1]{fontenc}
\usepackage[english]{babel}
\usepackage{mathptmx, amsmath, amssymb, amsfonts, amsthm, mathptmx, enumerate, color}
\usepackage[top=3.5cm, bottom=3.5cm, left=3.5cm, right=3.5cm]{geometry} 

\usepackage[utf8]{inputenc}
\usepackage[table]{xcolor}
\usepackage{graphicx}
\usepackage{amssymb,latexsym}
\usepackage{thmtools}
\usepackage{comment}
\usepackage{longtable}
\usepackage{enumerate}
\usepackage{hyperref}
\usepackage[ruled,vlined]{algorithm2e}  
\usepackage{float}
\usepackage{booktabs}
\usepackage{todonotes}
\usepackage{tabularx}
\usepackage{subcaption}
\usepackage{multirow}
\usepackage{etoolbox} 

\makeatletter
\patchcmd{\algocf@makecaption@ruled}{\hsize}{\textwidth}{}{} 
\patchcmd{\@algocf@start}{-1.5em}{0em}{}{} 
\makeatother

\DeclareMathOperator*{\argmin}{arg\,min}

\definecolor{lightsalmon}{rgb}{1.0, 0.63, 0.48}

\def\bs#1{\ensuremath{\boldsymbol{#1}}}

\theoremstyle{definition}

\newtheorem{remark}{Remark}[section]

\numberwithin{equation}{section}

\begin{document}
\title{Deep Generative Clustering with VAEs and Expectation-Maximization}
%
%
\author{Michael Adipoetra and Ségolène Martin \\
TU Berlin}
\date{ }
%
%
\maketitle              
%
\begin{abstract} We propose a novel deep clustering method that integrates Variational Autoencoders (VAEs) into the Expectation-Maximization (EM) framework. Our approach models the probability distribution of each cluster with a VAE and alternates between updating model parameters—by maximizing the Evidence Lower Bound (ELBO) of the log-likelihood—and refining cluster assignments based on the learned distributions. This enables effective clustering and generation of new samples from each cluster. Unlike existing VAE-based methods, our approach eliminates the need for a Gaussian Mixture Model (GMM) prior or additional regularization techniques. Experiments on MNIST and FashionMNIST demonstrate superior clustering performance compared to state-of-the-art methods.

\end{abstract}
\section{Introduction}
Clustering of images is a fundamental challenge in machine learning research \cite{jain1999data,xu2005survey}. While classical non-neural methods like K-means \cite{macqueen1967some} or the Expectation-Maximization (EM) algorithm with a Gaussian Mixture Model \cite{pearson1894contributions} are still widely used in many applications, these traditional approaches struggle to effectively capture the complex structures of data, such as image distributions. This limitation arises from the underlying assumption that the distribution of each cluster is Gaussian, centered around a parameter learned from the data. However, this assumption is often too simplistic.

To address these limitations, dimensionality reduction techniques can be used to project data into a lower-dimensional space where separation becomes easier. These transformations can be linear, such as Principal Component Analysis (PCA) \cite{pearson1901pca}, or nonlinear, using neural networks to learn complex data representations \cite{hinton2006oriNN,xie2016unsupervised}. When the network, often referred to as an \emph{encoder}, is learned independently of the clustering task, classical methods like K-means can still be applied on the latent representations. However, the results may be suboptimal since the encoding is not tailored for clustering on the given data. Designing an encoder specifically for clustering is challenging in the absence of labeled data.

In recent years, generative models have gained popularity for clustering tasks. These models enable the learning of the (multi-modal) data distribution using only unlabeled training data \cite{ghasedi2019balanced,mukherjee2019clustergan}. Interestingly, the use of a generative model for clustering goes beyond the clustering tasks and also opens up the possibility of generating new data points from the learned distribution, thus enriching classification datasets without the need of human labeling.

In particular, Variational Autoencoders (VAE) \cite{kingma2013auto,Kingma2019VAE}, which consist of an encoder that maps data to a low-dimensional latent space and a decoder that reconstructs the data, have been used as a means to simultaneously learn meaningful latent representations and cluster data. 
Several VAE-based clustering methods have been proposed in recent years. Variational Deep Embedding (VaDE) \cite{ijcai2017VaDE} employs a Gaussian Mixture Model (GMM) prior in the latent space with trainable parameters, initialized using a pretrained model. The Gaussian Mixture VAE (GMVAE) method \cite{dilokthanakul2016gmvae} offers a similar approach but eliminates the need for pretraining. However, GMVAE often suffers from cluster aggregation in the latent space, necessitating a cut-off trick during training to stabilize performance. To address such issues, some studies have introduced regularization terms in the loss function. For example, \cite{ma2022vaeloss} proposed a similarity-based loss as a graph embedding extension to ensure a more compact and well-separated latent space. Similarly to VaDE, \cite{prasad2020variational} introduced a model (referred to here as VAEIC) that avoids pretraining and incorporates an augmentation loss to maintain consistent cluster assignments between original and augmented data.

In this paper, we propose a novel VAE-based clustering method. Unlike existing methods such as VaDE, GMVAE, and VAEIC, our method eliminates the need for a GMM prior in the latent space and avoids any regularization strategies. Our method integrates VAE with an EM clustering framework, inspired by the few-shot classification method proposed in \cite{martin2022towards}. We formalize our method by deriving a lower bound for the expected log-likelihood, expressed as the Evidence Lower Bound (ELBO). The EM algorithm is then used to maximize the bound, where it alternately updates the clustering assignments and the VAE parameters for each cluster. 

The remainder of this paper is organized as follows. In Section \ref{sec_2}, we provide a brief overview of the EM algorithm and VAE. In Section \ref{sec_3}, we describe how our approach integrates the EM algorithm with the VAE generative framework for clustering. Finally, in Section \ref{sec_4}, we present experimental results, comparing our method's clustering performance with other VAE-based clustering methods and showcasing the generated samples.

\section{Background}\label{sec_2}

Our method leverages the EM algorithm to learn cluster assignments and VAE parameters, enabling sample generation from each cluster. Section \ref{sec_21} reviews the EM algorithm, and Section \ref{sec_22} covers Variational Autoencoders (VAEs), a key component of our approach.

\subsection{Expectation-Maximization Algorithm}\label{sec_21}

The Expectation-Maximization (EM) algorithm provides an iterative framework to maximize log-likelihood in the presence of latent variables. 
Let $\mathcal{X}$ denote the data space. Here, we assume $\mathcal{X} = \mathbb{R}^d$ and model the data using a random variable $X : \Omega \rightarrow \mathcal{X} = \mathbb{R}^d$ with unknown probability distribution $P_X$. We aim to find a distribution $P_{\bs{\alpha}}$ parameterized by model parameters $\bs{\alpha} \in \mathcal{A}$, where $\mathcal{A}$ represents the parameter space, such that $P_{\bs{\alpha}} \approx P_X$.
The corresponding density function (assumed to exist) is denoted as $p_{\bs{\alpha}}$. A common approach to determine the parameter $\bs{\alpha}$ is \textit{maximum likelihood estimation} (MLE) which involves maximizing the expected log-likelihood function, defined as
\begin{align*}
    \ell(\bs{\alpha}) := \mathbb{E}_{\bs{x} \sim P_X} [\log p_{\bs{\alpha}}(\bs{x})].
\end{align*}
MLE works effectively when the dataset is fully observed. However, the dataset might include hidden variables that are not directly observed, referred to as latent variables. These variables, denoted as $\bs{z}$, are realizations of a random variable $Z : \Omega \rightarrow \mathcal{Z}$ defined on some latent space $\mathcal{Z}$, which may differ from the data space $\mathcal{X}$. Although $\bs{z}$ is unobserved, it influences the modeling of $\bs{x}$. For example, in a mixture model, the latent variable $\bs{z}$ may represent an indicator that assigns data points to their respective clusters. Meanwhile, in VAE framework that we present next, the latent variable $\bs{z}$ is a low-dimensional latent representation that captures the underlying structure of the data $\bs{x}$. To account for its contribution, we marginalize over all possible values of $\bs{z}$, leading to the marginal log-likelihood
\begin{align*}
    \ell (\bs{\alpha}) = \log p_{\bs{\alpha}}(\bs{x}) = \log \int_z p_{\bs{\alpha}} (\bs{x},\bs{z}) \textnormal{d}\bs{z}.
\end{align*}
Maximizing this function over $\bs{\alpha}$ may be very difficult or impossible to do analytically due to the integral. To address this issue, the Expectation-Maximization (EM) algorithm was introduced in \cite{dempster1977EM}. 
Instead of directly maximizing the log-likelihood, it maximizes a lower bound on the log-likelihood, known as the Evidence Lower Bound (ELBO), which we re-derive for clarity. 
Using Bayes' formula, the joint density of observed data $\bs{x}$ and latent variable $\bs{z}$ is factorized as $p_{\bs{\alpha}}(\bs{x}, \bs{z}) = p_{\bs{\alpha}}(\bs{z}|\bs{x}) p_{\bs{\alpha}}(\bs{x})$, where $p_{\bs{\alpha}}(\bs{z}|\bs{x})$ is the posterior of $\bs{z}$ given $\bs{x}$. For any probability density function
\begin{align*}
  q_{\bs{x}} \in \mathcal{Q} := \left\{q : \mathcal{Z} \rightarrow [0,\infty), \int_{\bs{z} \in \mathcal{Z}} q(\bs{z}) d\bs{z} =  1\right\},
\end{align*}
the log-likelihood function can be decomposed as
\begin{align}\label{EM:elbo}
    \log(p_{\bs{\alpha}}(\bs{x})) &= \mathbb{E}_{\bs{z} \sim q_{\bs{x}}} \left[ \log \left( \frac{p_{\bs{\alpha}}(\bs{x},\bs{z})}{q_{\bs{x}}(\bs{z})} \right) \right] + \text{KL} \left( q_{\bs{x}}(\cdot) \| p_{\bs{\alpha}}(\cdot|\bs{x}) \right),
\end{align}
where $\textnormal{KL}(\cdot || \cdot)$ denotes the Kullback-Leibler divergence. Here, both the model parameter $\bs{\alpha}$ and the density $q_{\bs{x}}$ are unknown. This decomposition introduces the term Evidence Lower Bound (ELBO), a function of $\bs{\alpha}$ and $q_{\bs{x}}$, as
\begin{align}
    \textnormal{ELBO}(\bs{\alpha},q_{\bs{x}};\bs{x}) := \mathbb{E}_{\bs{z} \sim q_{\bs{x}}} \left[ \log \left( \frac{p_{\bs{\alpha}}(\bs{x},\bs{z})}{q_{\bs{x}}(\bs{z})} \right) \right].
\end{align}
Since KL is non-negative, ELBO provides a lower bound for $\log(p_{\bs{\alpha}}(\bs{x}))$ and the bound is tight if $q_{\bs{x}} = p_{\bs{\alpha}}(\cdot|\bs{x})$ for all $\bs{x}$. 

To use the EM algorithm, it is important to choose a model where the posterior distribution $p_{\bs{\alpha}}(\cdot|\bs{x})$ can be computed in a closed form (e.g., a Gaussian Mixture Model \cite{pearson1894contributions}).
In order to optimize ELBO, the EM algorithm alternates between updates on the density function $q_{\bs{x}} \in \mathcal{Q}$ and the model parameter $\bs{\alpha} \in \mathcal{A}$. For each iteration $j$, the following steps are performed.

\subsubsection{E-step.} Given the current parameter $\bs{\alpha}^{(j)}$, the density $q_{\bs{x}}$ is updated by maximizing $\textnormal{ELBO}(\bs{\alpha}^{(j)},q_{\bs{x}};\bs{x})$ with respect to $q_{\bs{x}}$.
It is equivalent to minimizing the KL divergence 
$
  \text{KL} \left( q_{\bs{x}} \| p_{\bs{\alpha}^{(j)}}(\cdot|\bs{x}) \right)$,
which achieves its minimum value of zero for $q_{\bs{x}} = p_{\bs{\alpha}^{(j)}}(\cdot|\bs{x}).$

\subsubsection{M-step.} In the maximization step, we aim to update $\bs{\alpha}$ while keeping $q$ fixed from the E-step. Using $q$ from the E-step (depending on $\bs{\alpha}^{(j)}$), ELBO can be decomposed as
\begin{align*}
    &\mathbb{E}_{\bs{z} \sim q_{\bs{x}}} \left[ \log \left( \frac{p_{\bs{\alpha}}(\bs{x},\bs{z})}{q_{\bs{x}}(\bs{z})} \right) \right] 
    = \underbrace{\mathbb{E}_{\bs{z} \sim q_{\bs{x}}} \left[ \log \left( p_{\bs{\alpha}}(\bs{x},\bs{z}) \right) \right]}_{=:U(\bs{\alpha}; \bs{\alpha}^{(j)}, \bs{x})} - \mathbb{E}_{\bs{z} \sim q_{\bs{x}}} \left[ \log \left( q_{\bs{x}}(\bs{z}) \right) \right].
\end{align*}
Since the second term only depends on $\bs{\alpha}^{(j)}$ and not on $\bs{\alpha}$, the M-step reduces to maximizing $\mathbb{E}_{\bs{x} \sim P_X} \left[U(\bs{\alpha}; \bs{\alpha}^{(j)}, \bs{x}) \right]$ , which can nevertheless be tedious.
Dempster, Laird and Rubin \cite{dempster1977EM} introduced the Generalized Expectation-Maximization (GEM) algorithm, which relaxes this requirement by only requiring that the updated parameter $\bs{\alpha}^{(j+1)}$ satisfies the increasing condition $$\mathbb{E}_{\bs{x} \sim P_X} \left[
    U(\bs{\alpha}^{(j+1)}; \bs{\alpha}^{(j)}, \bs{x}) \right] \geq \mathbb{E}_{\bs{x} \sim P_X} \left[U(\bs{\alpha}^{(j)}; \bs{\alpha}^{(j)}, \bs{x})\right].$$

Similarly to the EM algorithm, in the VAE framework, one optimizes a likelihood-based objective to learn the data distribution. However, while the EM algorithm assumes that the posterior distribution $p_{\bs{\alpha}}(\cdot|\bs{x})$ can be computed, VAE use neural networks to approximate it. 

\subsection{VAE}\label{sec_22}
Variational Autoencoders (VAEs), introduced by Kingma and Welling \cite{Kingma2019VAE}, are a generative modeling framework that incorporates a latent variable $Z : \Omega \to \mathcal{Z}$, where $\mathcal{Z} = \mathbb{R}^n$ represents a low-dimensional latent space ($n \ll d$). The latent variable $Z$ follows a simple prior distribution $P_Z$, typically a standard normal distribution, making sampling tractable even in higher dimensions.

A VAE approximates the data distribution $P_X$ using $P_Z$, employing a pair of stochastic mappings: the encoder and decoder, denoted by $(E, D)$. These mappings are used in the inference process (to estimate latent variable $\bs{z}$ from data $\bs{x}$) and the generation process (to reconstruct or generate $\bs{x}$ from $\bs{z}$), akin to dimensionality reduction techniques. The latent variable $\bs{z}$ captures a low-dimensional representation of the data, similar to the role of unobserved variables in the EM algorithm.


\subsubsection{Inference process.} Given an observed data $\bs{x}$, this process aims to estimate the latent variable $\bs{z}$. Specifically, for $\bs{x} \in \mathbb{R}^d$ drawn from the data distribution $P_X$, the neural network for the encoder $E = E_{\bs{\phi}} : \mathbb{R}^d \rightarrow \mathbb{R}^n \times \mathbb{R}^{n \times n}$ outputs the parameters $(\mu_{\bs{\phi}}(\bs{x}), \Sigma_{\bs{\phi}}(\bs{x}))$ of a multivariate Gaussian distribution $\mathcal{N}(\mu_{\bs{\phi}}(\bs{x}), \Sigma_{\bs{\phi}}(\bs{x}))$ with $
    E_{\bs{\phi}} (\bs{x}) := (\mu_{\bs{\phi}}(\bs{x}), \Sigma_{\bs{\phi}}(\bs{x}))$.
The density of the stochastic encoder $E_{\bs{\phi}}$, often called inference model is given by
\begin{equation*}
    q_{\bs{\phi}}(\bs{z}| \bs{x}) = \mathcal{N}(\bs{z};\mu_{\bs{\phi}}(\bs{x}), \Sigma_{\bs{\phi}}(\bs{x})).
\end{equation*}
In practice, the covariance $\Sigma_{\bs{\phi}}(\bs{x})$ is often assumed diagonal, with the encoder outputting the mean and log variance, i.e., $E_{\bs{\phi}} (\bs{x}) = (\mu_{\bs{\phi}}(\bs{x}), \log(\sigma^2_{\bs{\phi}}(\bs{x}))).$
This results in the density:
\begin{equation*}
    q_{\bs{\phi}}(\bs{z}| \bs{x}) = \mathcal{N}\left(\bs{z} ; \mu_{\bs{\phi}}(\bs{x}), \text{diag}(\sigma^2_{\bs{\phi}}(\bs{x}))\right).
\end{equation*}

\subsubsection{Generation process.} After obtaining the latent variable $\bs{z}$ from data $\bs{x}$ in the inference process, we can reconstruct $\bs{x}$ using $\bs{z}$ or generate new data by sampling $\bs{z}$. For reconstruction, $\bs{z} \sim q_{\bs{\phi}}(\cdot | \bs{x})$, and for random generation, $\bs{z} \sim \mathcal{N}(0, I)$. The decoder neural network $D_{\bs{\theta}} : \mathbb{R}^n \rightarrow \mathbb{R}^d$ maps $\bs{z}$ to the mean $\mu_{\bs{\theta}}(\bs{z})$ of a multivariate Gaussian distribution with identity covariance, where $
    D_{\bs{\theta}} (\bs{z}) := \mu_{\bs{\theta}}(\bs{z})$,
and the density of the decoder, known as generation model, is
\begin{equation}
    p_{\bs{\theta}}(\bs{x} | \bs{z}) = \mathcal{N}(\bs{x};\mu_{\bs{\theta}}(\bs{z}), I).
\end{equation}
The reconstructed or generated data $\bs{x}$ is obtained as $
    \bs{x} = \mu_{\bs{\theta}}(\bs{z})$.

\subsubsection{Training.} In the following, we denote $\bs{\alpha} := (\bs{\phi},\bs{\theta})$, where $\bs{\phi}$ and $\bs{\theta}$ are the neural network parameters for encoder and decoder of the VAE respectively. In order to learn these parameters $\bs{\alpha}$, Kingma \cite{Kingma2019VAE} seeks to maximizing the log-likelihood $\mathbb{E}_{\bs{x} \sim P_X} [\log p_{\bs{\theta}}(\bs{x})]$ by optimizing ELBO, which can further be derived as
\begin{align}\label{eq:ELBO_training}
    \textnormal{ELBO}(\bs{\alpha};\bs{x}) =  \mathbb{E}_{\bs{z} \sim q_{\bs{\phi}}(\cdot|\bs{x})} \left[ \log p_{\bs{\theta}}(\bs{x}|\bs{z}) 
    \right] - \textnormal{KL}(q_{\bs{\phi}}(\bs{z}|\bs{x}) || p_Z(\bs{z})). 
\end{align}
For a detailed derivation of the ELBO, see \cite{gaby2023SNF_2}. The first term, known as reconstruction loss, encourages the decoder to reconstruct $\bs{x}$ accurately for a given $\bs{z}$. This is expressed as
\begin{align*}
    \mathbb{E}_{\bs{z} \sim q_{\bs{\phi}}(\cdot|\bs{x})} \left[ \log p_{\bs{\theta}}(\bs{x}|\bs{z}) 
    \right] = - \frac{1}{2}\mathbb{E}_{\bs{z} \sim \mathcal{N}\left( [E_{\bs{\phi}}(\bs{x})]_1,\text{diag}(\exp[E_{\bs{\phi}}(\bs{x})]_2)\right)} \left[ ||\bs{x} - D_{\bs{\theta}} (\bs{z})||^2\right].
\end{align*}
The second term regularizes the inference model to align with the prior. The KL divergence between two multivariate Gaussians has the closed-form expression
\begin{align*}
    \textnormal{KL}(q_{\bs{\phi}}(\bs{z}|\bs{x}) || p_Z(\bs{z})) &= \frac{1}{2} \sum_{i=1}^n  \left( - 1 - [[E_{\bs{\phi}} (\bs{x})]_2]_i + [\exp([E_{\bs{\phi}} (\bs{x})]_2)]_i + [[E_{\bs{\phi}} (\bs{x})]_1]_i^2\right).
\end{align*}
However, the gradient of the ELBO with respect to the parameter $\bs{\phi}$ is intractable, so we use the \emph{reparameterization trick} introduced in \cite{Kingma2019VAE}, enabling computation of the reconstruction loss as:
\begin{align*}
    &\mathbb{E}_{\bs{z} \sim q_{\bs{\phi}}(\cdot|\bs{x})} \left[ \log p_{\bs{\theta}}(\bs{x}|\bs{z}) 
    \right] \\
    = &- \frac{1}{2}\mathbb{E}_{\bs{\varepsilon} \sim \mathcal{N}(0,I)} \left[ \Vert x - D_{\bs{\theta}} \left([E_{\bs{\phi}}(\bs{x})]_1 +  \exp([E_{\bs{\phi}}(\bs{x})]_2 /2)\odot \bs{\varepsilon}\right)\Vert^2\right].
\end{align*}

\section{Proposed Approach}\label{sec_3}

We assume access to $N$ data points $\bs{x}_1, \dots, \bs{x}_N$ from $K \in \mathbb{N}$ distinct clusters, with $K$ assumed to be known. The objective is twofold: (1) to partition the data into $K$ clusters without access to labels, and (2) to learn the distribution of each cluster. At inference, the learned distributions can be leveraged to (i) assign test samples to a cluster and (ii) generate new samples from each cluster. 

Note that training a generic generative model (e.g., VAE, GAN, flow-based model) on a multi-class dataset allows for generating new samples from the overall dataset but not from specific classes. A common approach to address this is to condition the model on class labels \cite{mirza2014conditional}. In contrast, our method enables generation of samples from specific clusters without requiring labels.

Our method builds on the transductive few-shot clustering algorithm introduced by Martin et al.~\cite{martin2022towards} and extended in~\cite{martin2024transductive}. While their method addresses a different problem—few-shot classification, where the goal is to classify with very limited training samples—it shares key methodological similarities with ours. In these works, a pretrained encoder (trained on a generic dataset) is used to obtain low-dimensional representations of images. The distributions of the clusters in this latent space are then modeled using a parametric distribution, such as Dirichlet distributions. Their algorithm alternates between optimizing cluster assignments and estimating the parameters of these cluster distributions.
In contrast, while we adopt a similar alternating optimization strategy, our approach learns the distribution of each cluster directly in the data space using a generative model. This enables us to avoid dependence on a fixed latent representation.

Assume that each data point $\bs{x}$ is generated from a mixture model~\cite{pearson1894contributions} with fixed mixing coefficients $\bs{\pi} = \left(\frac{1}{K}, \ldots, \frac{1}{K}\right)$. The probability density function of the mixture model is given by:
\begin{align}\label{mixture_model}
    p_{\bs{\alpha}}(\bs{x}) = \sum_{k=1}^K \pi_k p_{\bs{\alpha}_k}(\bs{\bs{x}}),
\end{align}
where $\pi_k = \frac{1}{K}$ denotes the mixing coefficient for cluster $k$, $p_{\bs{\alpha}_k}(\bs{x})$ is the probability density function of cluster $k$, and $\bs{\alpha}_k$ represents the parameters specific to that cluster.
The expected log-likelihood corresponding to the mixture model is:
\begin{align}\label{expected_log_likelihood}
    \mathbb{E}_{\bs{x} \sim P_X}[\log(p_{\bs{\alpha}}(\bs{x}))] 
    &= \sum_{i=1}^N \log(p_{\bs{\alpha}}(\bs{x}_i))
    = \sum_{i=1}^N \log \left( \sum_{k=1}^K \frac{1}{K} p_{\bs{\alpha}_k}(\bs{x}_i) \right).
\end{align}


Futhermore, we introduce for each data point $\bs{x}_i$ a soft assignment variable $\bs{u}_i \in \Delta_K$, where $\Delta_K$ denotes the probability simplex over $K$ classes:
\begin{align*}
    \Delta_K = \left\{ \bs{u} = (u_k)_{1 \leq k \leq K} \in \mathbb{R}^K \ \middle|\ u_k \geq 0,\ \forall k \in \{1, \dots, K\},\ \text{and} \ \sum_{k=1}^{K} u_k = 1 \right\}.
\end{align*}
Each element $u_{i,k}$ in the vector $\bs{u}_i$ represents the probability that sample $\bs{x}_i$ belongs to cluster $k$. 
Leveraging the concavity of the $\log$ function and starting from Eq.~\eqref{mixture_model}, we use the property that $\bs{u}_i \in \Delta_K$ to derive a lower bound for the expected log-likelihood:
\begin{align*}
    \mathbb{E}_{\bs{x} \sim P_X}[\log(p_{\bs{\alpha}}(\bs{x}))] 
    &\simeq \sum_{i=1}^N \log \left(\sum_{k=1}^K u_{i,k}\frac{p_{\bs{\alpha}_k}(\bs{x}_i)}{K u_{i,k}}\right) \\
    &\geq \sum_{i=1}^N \sum_{k=1}^K u_{i,k}\log \left(\frac{p_{\bs{\alpha}_k}(\bs{x}_i)}{K u_{i,k}}\right) \\
    &= \sum_{i=1}^N \sum_{k=1}^K u_{i,k}\log (p_{\bs{\alpha}_k}(\bs{x}_i)) - \sum_{i=1}^N \sum_{k=1}^K u_{i,k}\log (u_{i,k}) - N \log (K).
\end{align*}
This lower bound of the log-likelihood leads to the optimization problem:
\begin{alignat}{2}\label{clustering}
    &\underset{(\bs{u}_i)_{1 \leq i \leq N}, (\bs{\alpha}_k)_{1 \leq k \leq K}}{\mathrm{minimize}} 
    &\quad & -\sum_{i=1}^N \sum_{k=1}^K u_{i,k} \log p_{\bs{\alpha}_k}(\bs{x}_i) + \sum_{i=1}^N \sum_{k=1}^K u_{i,k} \log u_{i,k}, \\
    &\qquad \text{subject to} 
    &\quad & \bs{u}_i \in \Delta_K \quad \forall i \in \{1, \ldots, N\}.\nonumber
\end{alignat}

\begin{remark}
The minimization problem \eqref{clustering} aligns with the EM framework described in Section \ref{sec_21}. In this context, the latent space is the discrete space of clusters, i.e., $\mathcal{Z} = \{1, \dots, K\}$. The random variable $Z$ is the uniform cluster assignment, and for all $i \in \{1, \dots, N\}$, the approximate posterior $q_{\bs{x}_i}$ is defined as $
q_{\bs{x}_i} = \sum_{k=1}^K u_{i,k} \delta_k$,
where $\delta_k$ is the Dirac delta function centered at $k$.
The Expectation (E) step involves minimizing the following expression:
\begin{align*}
    - \sum_{i=1}^N \mathbb{E}_{z \sim q_{\bs{x}_i}}\left[ \log\left(\frac{p_{\bs{\alpha}}(z | \bs{x}_i)}{q_{\bs{x}_i}(z)} \right) \right] 
    &= -\sum_{i=1}^N \sum_{k=1}^K u_{i,k} \log\left(\frac{p_{\bs{\alpha}}(z= k | \bs{x}_i)}{u_{i,k}} \right),\\
    &= -\sum_{i=1}^N \sum_{k=1}^K u_{i,k} \log\left(\frac{p_{\bs{\alpha}}(\bs{x}_i|z=k)}{p_{\bs{\alpha}}(\bs{x}_i)u_{i,k}} \frac{1}{K} \right) ,\\
    &= -\sum_{i=1}^N \sum_{k=1}^K u_{i,k} \left(\log\left(\frac{p_{\bs{\alpha_k}}(\bs{x}_i)}{Ku_{i,k}}\right)\right) + c,
\end{align*}
where $c$ is a constant independent on $(\bs{u}_{i})_{1\leq i \leq N}$. We thus retrieve \eqref{clustering}.
\end{remark}
\
We now propose modeling the log-probability function $\log(p_{\bs{\alpha}_k}(\cdot))$ for each cluster $k \in \{1, \dots, K\}$ using $K$ distinct VAEs (one per cluster) with encoder and decoder parameters $\bs{\alpha}_k = (\bs{\phi}_k, \bs{\theta}_k)$, as defined in Section \ref{sec_22}. Leveraging the ELBO from \eqref{eq:ELBO_training}, the minimization problem \eqref{clustering} simplifies to:
\begin{alignat}{2}\label{TGZS}
    &\underset{(\bs{u}_i)_{1 \leq i \leq N}, (\bs{\alpha}_k)_{1 \leq k \leq K}}{\mathrm{minimize}}  &\quad &- \sum_{i=1}^N\sum_{k=1}^K u_{i,k} \textnormal{ELBO}(\bs{\alpha}_k;\bs{x}_i) + \sum_{i=1}^N \sum_{k=1}^K u_{i,k} \log u_{i,k}, \\
     &\qquad \text{subject to} 
    &\quad & \bs{u}_i \in \Delta_K \quad \forall i \in \{1, \ldots, N\}. \nonumber
\end{alignat}
This optimization is performed iteratively via the EM algorithm.

\paragraph{Expectation Step.}

In the expectation step, we assume that the data $\bs{x}_i$ and the parameters $\bs{\alpha}_k$ are fixed. 
It is straightforward to observe that the minimization of \eqref{TGZS} with respect to $\bs{u}_i$ is equivalent to minimizing a Kullback-Leibler divergence over the simplex. The solution to this problem has a known closed-form expression \cite{benamou2015iterative}, 
given by:
\begin{equation*}
    \bs{u}_{i} = \text{softmax}\Big(\left\{\textnormal{ELBO}(\bs{\alpha}_k^{(j)};\bs{x}_i)\right\}_{1 \leq k \leq K}\Big)
\end{equation*}

\paragraph{Maximization Step.}
In the maximization step, the variables $\bs{u}_i$ are fixed, and the focus shifts to optimizing the parameters $\bs{\alpha}_k$ for each class $k \in \{1, \ldots, K\}$. Since the second term in the objective function does not depend on the neural network parameters, the optimization for each class simplifies to:
\begin{align*}
    \argmin_{\bs{\alpha}_k} - \sum_{i=1}^N u_{i,k} \textnormal{ELBO}(\bs{\alpha}_k;\bs{x}_i).
\end{align*}
The parameters $\bs{\alpha}_k$ are updated using the ADAM optimizer \cite{kingma2014adam}. As exact minimization of the ELBO is not guaranteed with gradient-based approaches, this procedure aligns with the Generalized Expectation-Maximization framework, which requires only that the ELBO decreases at each iteration.
The detailed process for solving the minimization problem in \eqref{TGZS} is presented in Algorithm \ref{alg:vae_clustering}.

\setlength{\algomargin}{0pt} 
\SetAlCapHSkip{0em}
\begin{algorithm}[H]
\SetKwInOut{Input}{Input}
\SetKwInOut{Output}{Output}
\SetKwInOut{Init}{Init}
\textbf{Input: }{$\{\bs{x}_1, \dots, \bs{x}_N\}$ train data, VAE models $(E_{\bs{\phi}_k}, D_{\bs{\theta}_k})$, for $k \in \{1, \dots, K\}$.\\
\textbf{Output: }{Assignments $(u_{i,k})_{1 \leq i \leq N \atop 1 \leq i \leq K}$ and model weights $\bs{\alpha} = \left\{ (\bs{\phi}_k, \bs{\theta}_k)\right\}_{1 \leq k \leq K}$.}
\caption{EM-Based Clustering with VAE}
\label{alg:vae_clustering}
\textbf{Init: }{Initialize soft assignments $\bs{u}_i \in \Delta_K$ randomly, initialize model weights $\bs{\alpha}_k^{(0)} = (\bs{\phi}_k^{(0)}, \bs{\theta}_k^{(0)})$}, $k \in \{1, \dots, K\}$, randomly.}\\
\For{$j = 1, 2, \dots, $ }{
    \textbf{M-Step:} Update VAE parameters for each cluster $k \in \{1, \dots, K\}$\;
     \[
        \bs{\alpha}_k^{(j)} = \argmin_{\bs{\alpha}_k} - \sum_{i=1}^N u_{i,k}^{(j-1)} \textnormal{ELBO}(\bs{\alpha}_k;\bs{x}_i)
    \]
    \textbf{E-Step:} Update soft assignments for $i \in \{1, \dots, N\}$; 
    \[
        \bs{u}_{i}^{(j)} = \text{softmax}\Big(\left\{\textnormal{ELBO}(\bs{\alpha}_k^{(j)};\bs{x}_i)\right\}_{1 \leq k \leq K}\Big)
    \]
}
\end{algorithm}

\section{Numerical experiments}\label{sec_4}

In this section, we present a comparative analysis of our proposed clustering method against several VAE-based clustering models. 

\subsection{Experimental setup}

\paragraph{Datasets.} 
We created a synthetic 2D dataset with 5000 points derived from arcs of 5 half-moon shapes. Additionally, we used the MNIST \cite{lecun1998mnist} and FashionMNIST \cite{xiao2017fashion} datasets, each comprising 70,000 grayscale images of $28 \times 28$ pixels. MNIST features handwritten digits, while FashionMNIST consists of fashion items.

\paragraph{Baselines.}
We compare our approach with clustering VAE-based methods, namely VaDE \cite{ijcai2017VaDE}, GMVAE \cite{dilokthanakul2016gmvae}, and VAEIC \cite{prasad2020variational}. Other clustering approaches, such as hierarchical VAE (e.g., LadderVAE \cite{sonderby2016ladder}, TreeVAE\cite{manduchi2023tree}), GAN-based methods like ClusterGAN \cite{mukherjee2019clustergan}, also exist and have demonstrated strong performance in related tasks. However, in this experiment, we focus solely on comparisons with VAE-based methods.
For VaDE, which relies on pretraining, we also report the accuracy of the pretraining step. The pretraining follows the original approach, using an autoencoder to reduce the data dimensionality to 10, followed by clustering with a Gaussian Mixture Model (GMM). This pretraining method is referred to as AE+GMM. GMVAE is implemented using 10 Monte Carlo (MC) samples and we set the number of clusters to be 10. All baseline methods are implemented using the architectures and hyperparameters proposed in their respective papers. 

\paragraph{Hardware.}
All tests were run on a single NVIDIA GeForce RTX 2080 Ti with 16 GB of memory.

\paragraph{Training of our method.} Our implementation uses a fully connected neural network architecture for both the encoder and decoder. The implementation details for our method
are summarized in Table \ref{tab:implementation_details}.
\begin{table}[htb]
\centering
\caption{Implementation Details for MNIST, FashionMNIST, and Synthetic Dataset. We put a '$-$' if we do not use the component.}
\label{tab:implementation_details}

\begin{tabular}{l|l|l}
\toprule
\textbf{Component}          & \textbf{MNIST and FashionMNIST} & \textbf{Synthetic Dataset} \\ \midrule
\textbf{Network Architecture} & Fully connected & Fully connected\\ 
\textbf{Encoder Dimensions}   & $784-500-20$ & $2-100-2$\\ 
\textbf{Decoder Dimensions}   & $20-500-784$ & $2-100-2$\\ 
\textbf{Activation Function}  & Leaky ReLU (slope 0.2) & Leaky ReLU (slope 0.2)\\ 
\textbf{Dropout Rate}         & 0.2 & $-$\\ 
\textbf{Learning Rate}        & 0.001 & 0.0001\\ 
\textbf{Decay}                & 0.9 every 20 EM iterations & $-$\\ 
\textbf{Optimizer}            & Adam \cite{kingma2014adam} & Adam \cite{kingma2014adam}\\ 
\textbf{Batch Size}           & 256 & 256\\
\textbf{Number MC samples (E-step)}           & 10 & 10\\
\textbf{Number MC samples (M-step)}           & 1 & 1\\
\textbf{Reconstruction loss}  & 1 & 5\\ 
\textbf{Training Epochs (M-step)} & 20 per iteration & 5 per iteration\\ 
\textbf{Total EM Iterations}  & 300 & 1000\\
\bottomrule
\end{tabular}
\end{table}

\subsection{Clustering Results}

Table \ref{tab:acc} shows the average and best clustering accuracy over 10 random runs for the MNIST and FashionMNIST datasets. AE+GMM performs worse, highlighting the effectiveness of VAE-based models. For MNIST, our method achieves the highest accuracy on both average and best runs. On FashionMNIST, all methods perform similarly, with GMVAE achieving the best single-run accuracy, while our approach leads in average accuracy, demonstrating greater consistency.
\begin{table}[htb]
    \centering
    \caption{Clustering accuracy $(\%)$. Best is shown in boldface and second best is underlined.}
    \begin{tabular}{l | c c | c c | c c | c c}
        \toprule
        \multicolumn{1}{c}{} & \multicolumn{2}{c}{\textit{\textbf{MNIST}-{train}}} & \multicolumn{2}{c}{\textit{\textbf{MNIST}-test}}  & \multicolumn{2}{c}{{\textit{\textbf{Fashion}-train}}} & \multicolumn{2}{c}{{\textit{\textbf{Fashion}-test}}} \\
        \cmidrule(rl){2-3} \cmidrule(rl){4-5} \cmidrule(rl){6-7} \cmidrule(rl){8-9} 
        Method & {Avg} & {Best} & {Avg}  & {Best} & {Avg}  & {Best} & {Avg}  & {Best} \\
        \midrule
        {GMVAE}  & $85.4\scriptstyle{\pm 4.1}$ & 93.1 &  $85.4\scriptstyle{\pm 4.1}$ & 93.54 & $\underline{58.3}\scriptstyle{\pm 5.8}$& \textbf{70.1} &  $\underline{58.1}\scriptstyle{\pm 5.6}$ & \textbf{69.2} \\
        VAEIC  & $86.8\scriptstyle{\pm 4.9}$ & 93.2 & $87.1 \scriptstyle{\pm 4.8}$ &  93.7  & $57.6 \scriptstyle{\pm 4.0}$ & \underline{63.6} & $57.3 \scriptstyle{\pm 3.9}$ & 63.5 \\
        AE+GMM  & $78.2\scriptstyle{\pm 1.7}$ & 81.4 & $78.6\scriptstyle{\pm 2.5}$   & 82.5 & $55.5 \scriptstyle{\pm 1.4}$ & 56.8 & $56.4 \scriptstyle{\pm 2.2}$ & 59.8 \\
        {VaDE}  & $\underline{88.1}\scriptstyle{\pm 2.9}$ & \underline{93.4} & $\underline{88.3}\scriptstyle{\pm 3.0}$ & \underline{94.2} & $57.2 \scriptstyle{\pm 1.6}$ & 59.4 & $57.0 \scriptstyle{\pm 1.7}$ & 59.8 \\
        \rowcolor{lightsalmon!20} Ours & $\textbf{88.4}\scriptstyle{\pm 5.2}$& \textbf{94.3} &  $\textbf{88.5}\scriptstyle{\pm 5.1}$ & \textbf{94.6} & $\textbf{59.1}\scriptstyle{\pm 2.9}$ & 63.3 &  $\textbf{58.9}\scriptstyle{\pm 3.2}$ & \underline{63.6} \\
        \bottomrule
    \end{tabular}
    \label{tab:acc}
\end{table}

\subsection{Generated Samples}

\begin{figure}[htb]
  \begin{subfigure}[b]{0.25\linewidth}
    \includegraphics[width=1\linewidth]{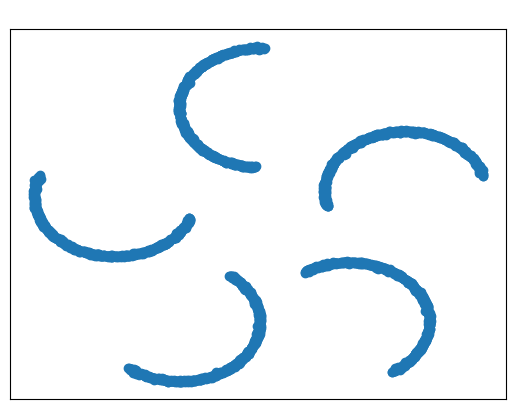} 
    \caption{Training samples} 
    \label{syn:train} 
  \end{subfigure}
  \begin{subfigure}[b]{0.25\linewidth}
    \includegraphics[width=1\linewidth]{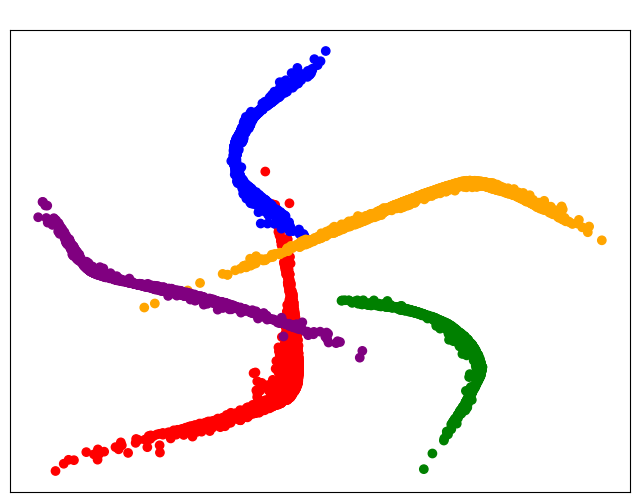} 
    \caption{Iter 100} 
    \label{syn:iter100} 
  \end{subfigure}
  \begin{subfigure}[b]{0.25\linewidth}
    \includegraphics[width=1\linewidth]{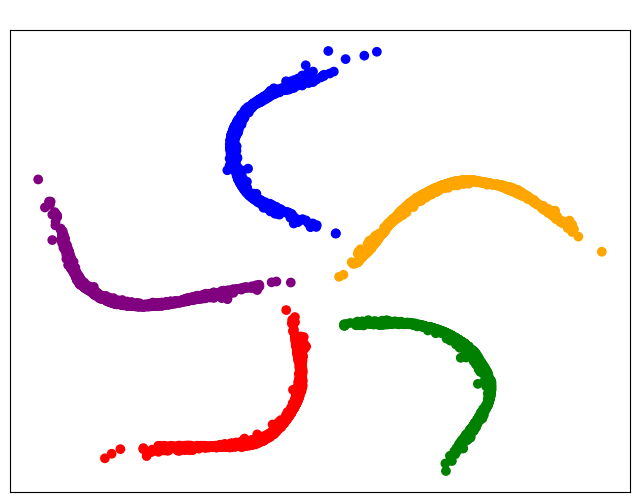} 
    \caption{Iter 200} 
    \label{syn:iter200} 
  \end{subfigure}
  \begin{subfigure}[b]{0.25\linewidth}
    \includegraphics[width=1\linewidth]{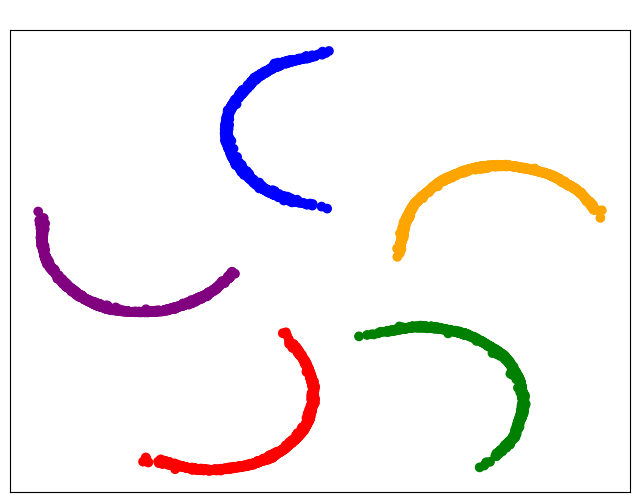} 
    \caption{Iter 1000} 
    \label{syn:iter1000} 
  \end{subfigure}
  \caption{Training samples (unlabeled) and generated samples after 100, 200, and 1000 EM iterations. Each color represents samples from the same cluster-specific VAE. Our method effectively clusters and generates samples.
  \label{fig:generated_moons}}
\end{figure}

\begin{figure}[htb] 
\centering
    \includegraphics[scale=0.1]{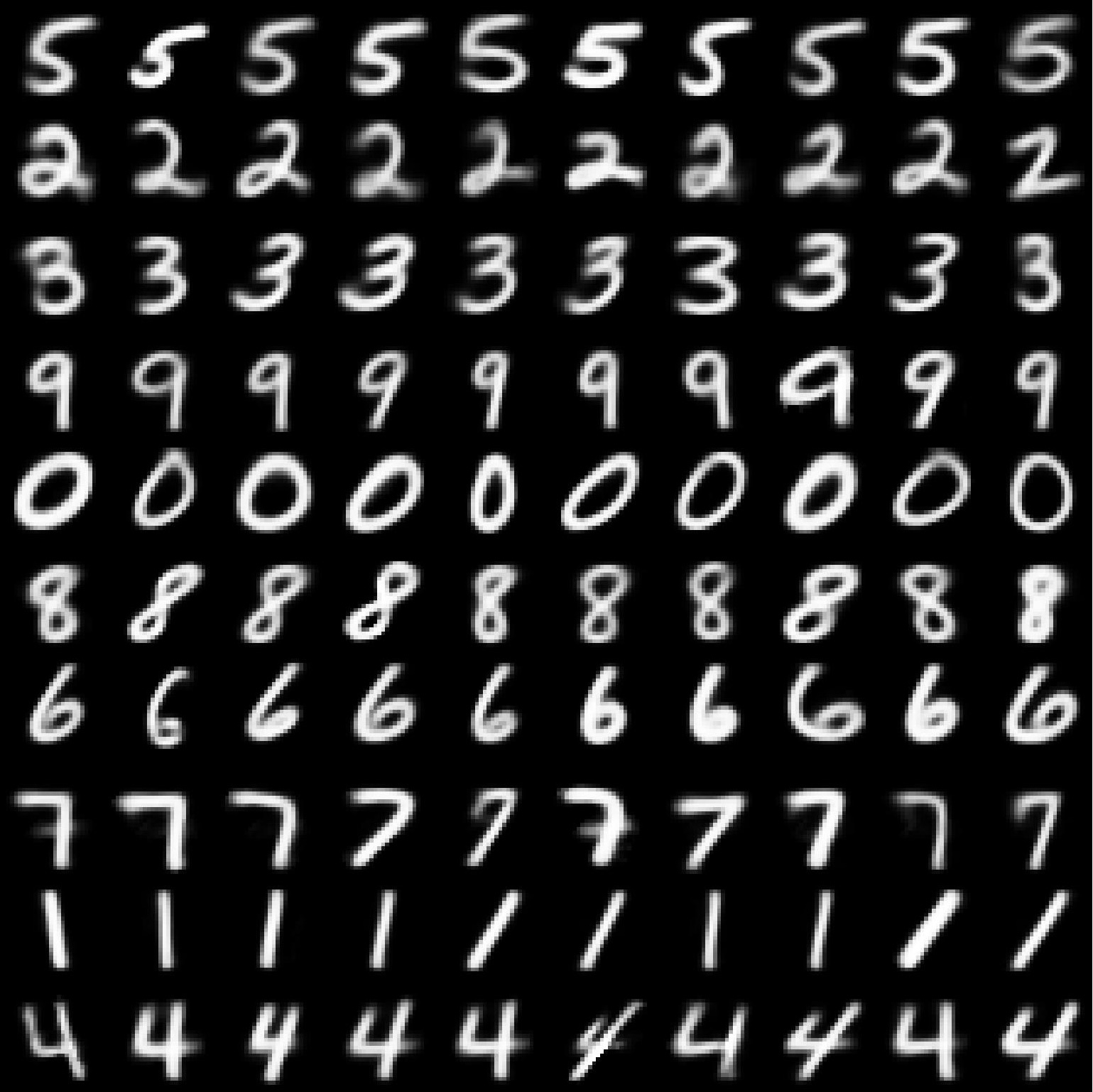}     \hspace{1cm}
    \includegraphics[scale=0.1]{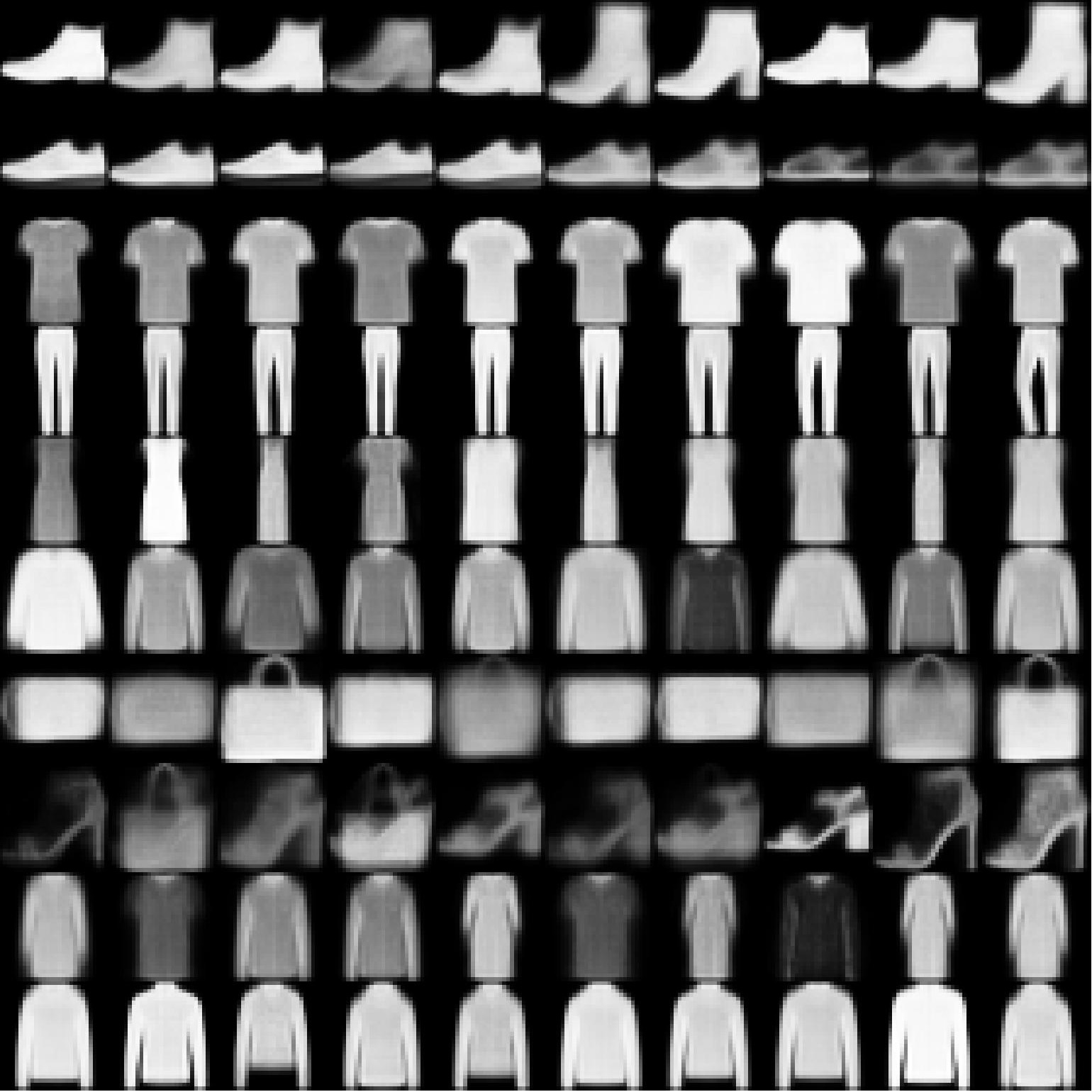} 
  \caption{Generated samples with our method on MNIST (left) and Fashion-MNIST (right). Images in the same row come from the same cluster.\label{fig:generated}}
\end{figure}

One key advantage of our proposed model is its ability to independently generate samples from each cluster. To generate new samples, we first draw random samples from the standard Gaussian distribution and pass them through the decoder of the corresponding cluster-specific VAE.
In Figures \ref{fig:generated_moons} and \ref{fig:generated}, we visualize the generated images from each cluster across the three datasets. We observe that: 1) our method effectively samples from each cluster independently with few class mismatch; 2) the generated samples exhibit high variability, and as shown in Figure \ref{fig:generated_moons}, our method covers the entire distribution of each cluster.


\section{Conclusion}
In this paper, we introduced a novel clustering framework that integrates VAEs with the EM algorithm for unsupervised clustering. Our approach alternately optimizes soft cluster assignments and VAE parameters for each cluster, enabling both effective clustering and the generation of new samples from each cluster using the respective VAE models. Empirical evaluations on synthetic and real datasets demonstrate that our model outperforms existing VAE-based clustering methods in terms of average clustering accuracy. Future work could extend this generative EM framework to other models, such as normalizing flows, which allow for easy computation of log-likelihoods without relying on lower bounds.

\vskip 6mm
\section*{Acknowledgments}

\noindent  S\'egol\`ene Martin was supported by the DFG funded Cluster of Excellence EXC 2046 MATH+ (project ID: AA5-8).


\bibliographystyle{abbrv}
\bibliography{bibliography.bib}

\begin{thebibliography}{10}

\bibitem{benamou2015iterative}
J.-D. Benamou, G.~Carlier, M.~Cuturi, L.~Nenna, and G.~Peyr{\'e}.
\newblock Iterative {B}regman projections for regularized transportation
  problems.
\newblock {\em SIAM Journal on Scientific Computing}, 37(2):A1111--A1138, 2015.

\bibitem{dempster1977EM}
A.~P. Dempster, N.~M. Laird, and D.~B. Rubin.
\newblock Maximum likelihood from incomplete data via the {EM} algorithm.
\newblock {\em Journal of the Royal Statistical Society: Series B},
  39(1):1--22, 1977.

\bibitem{dilokthanakul2016gmvae}
N.~Dilokthanakul, P.~A. Mediano, M.~Garnelo, M.~C. Lee, H.~Salimbeni,
  K.~Arulkumaran, and M.~Shanahan.
\newblock Deep unsupervised clustering with {G}aussian mixture variational
  autoencoders.
\newblock {\em arXiv preprint arXiv:1611.02648}, 2016.

\bibitem{ghasedi2019balanced}
K.~Ghasedi, X.~Wang, C.~Deng, and H.~Huang.
\newblock Balanced self-paced learning for generative adversarial clustering
  network.
\newblock In {\em IEEE/CVF Conference on Computer Vision and Pattern
  Recognition}, pages 4391--4400, 2019.

\bibitem{gaby2023SNF_2}
P.~L. Hagemann, J.~Hertrich, and G.~Steidl.
\newblock Generalized {N}ormalizing {F}lows via {M}arkov {C}hains.
\newblock In {\em Elements in Non-local Data Interactions: Foundations and
  Applications.} Cambridge University Press, 2023.

\bibitem{hinton2006oriNN}
G.~E. Hinton and R.~R. Salakhutdinov.
\newblock Reducing the dimensionality of data with neural networks.
\newblock {\em science}, 313(5786):504--507, 2006.

\bibitem{jain1999data}
A.~K. Jain, M.~N. Murty, and P.~J. Flynn.
\newblock Data clustering: a review.
\newblock {\em ACM Computing Surveys}, 31(3):264--323, 1999.

\bibitem{ijcai2017VaDE}
Z.~Jiang, Y.~Zheng, H.~Tan, B.~Tang, and H.~Zhou.
\newblock Variational deep embedding: An unsupervised and generative approach
  to clustering.
\newblock In {\em International Joint Conference on Artificial Intelligence,
  (IJCAI)}, pages 1965--1972, 2017.

\bibitem{kingma2014adam}
D.~P. Kingma.
\newblock {ADAM}: A method for stochastic optimization.
\newblock {\em arXiv preprint arXiv:1412.6980}, 2014.

\bibitem{kingma2013auto}
D.~P. Kingma.
\newblock Auto-encoding variational bayes.
\newblock In {\em International Conference on Learning Representations}, 2014.

\bibitem{Kingma2019VAE}
D.~P. Kingma and M.~Welling.
\newblock An introduction to variational autoencoders.
\newblock {\em Foundations and Trends in Machine Learning}, 12(4):307–392,
  2019.

\bibitem{lecun1998mnist}
Y.~LeCun, L.~Bottou, Y.~Bengio, and P.~Haffner.
\newblock Gradient-based learning applied to document recognition.
\newblock {\em Proceedings of the IEEE}, 86(11):2278--2324, 1998.

\bibitem{ma2022vaeloss}
H.~Ma.
\newblock Achieving deep clustering through the use of variational autoencoders
  and similarity-based loss.
\newblock {\em Mathematical Biosciences and Engineering}, 19(10):10344--10360,
  2022.

\bibitem{macqueen1967some}
J.~MacQueen.
\newblock Some methods for classification and analysis of multivariate
  observations.
\newblock In {\em Berkeley Symposium on Mathematical Statistics and
  Probability}, 1967.

\bibitem{manduchi2023tree}
L.~Manduchi, M.~Vandenhirtz, A.~Ryser, and J.~Vogt.
\newblock Tree variational autoencoders.
\newblock {\em Advances in Neural Information Processing Systems},
  36:54952--54986, 2023.

\bibitem{martin2022towards}
S.~Martin, M.~Boudiaf, E.~Chouzenoux, J.-C. Pesquet, and I.~Ayed.
\newblock Towards practical few-shot query sets: transductive minimum
  description length inference.
\newblock {\em Advances in Neural Information Processing Systems},
  35:34677--34688, 2022.

\bibitem{martin2024transductive}
S.~Martin, Y.~Huang, F.~Shakeri, J.-C. Pesquet, and I.~Ben~Ayed.
\newblock Transductive zero-shot and few-shot {CLIP}.
\newblock In {\em IEEE/CVF Conference on Computer Vision and Pattern
  Recognition}, pages 28816--28826, 2024.

\bibitem{mirza2014conditional}
M.~Mirza.
\newblock Conditional generative adversarial nets.
\newblock {\em arXiv preprint arXiv:1411.1784}, 2014.

\bibitem{mukherjee2019clustergan}
S.~Mukherjee, H.~Asnani, E.~Lin, and S.~Kannan.
\newblock Cluster{GAN}: Latent space clustering in generative adversarial
  networks.
\newblock In {\em Proceedings of the AAAI Conference on Artificial
  Intelligence}, volume~33, pages 4610--4617, 2019.

\bibitem{pearson1894contributions}
K.~Pearson.
\newblock Contributions to the mathematical theory of evolution.
\newblock {\em Philosophical Transactions of the Royal Society of London. A},
  185:71--110, 1894.

\bibitem{pearson1901pca}
K.~Pearson.
\newblock On lines and planes of closest fit to systems of points in space.
\newblock {\em The London, Edinburgh, and Dublin Philosophical Magazine and
  Journal of Science}, 2(11):559--572, 1901.

\bibitem{prasad2020variational}
V.~Prasad, D.~Das, and B.~Bhowmick.
\newblock Variational clustering: Leveraging variational autoencoders for image
  clustering.
\newblock In {\em International Joint Conference on Neural Networks}, pages
  1--10. IEEE, 2020.

\bibitem{sonderby2016ladder}
C.~K. S{\o}nderby, T.~Raiko, L.~Maal{\o}e, S.~K. S{\o}nderby, and O.~Winther.
\newblock Ladder variational autoencoders.
\newblock {\em Advances in Neural Information Processing Systems}, 29, 2016.

\bibitem{xiao2017fashion}
H.~Xiao, K.~Rasul, and R.~Vollgraf.
\newblock Fashion-{MNIST}: a novel image dataset for benchmarking machine
  learning algorithms.
\newblock {\em arXiv preprint arXiv:1708.07747}, 2017.

\bibitem{xie2016unsupervised}
J.~Xie, R.~Girshick, and A.~Farhadi.
\newblock Unsupervised deep embedding for clustering analysis.
\newblock In {\em International Conference on Machine Learning}, pages
  478--487. PMLR, 2016.

\bibitem{xu2005survey}
R.~Xu and D.~Wunsch.
\newblock Survey of clustering algorithms.
\newblock {\em IEEE Transactions on Neural Networks}, 16(3):645--678, 2005.

\end{thebibliography}

\end{document}